\documentclass[letterpaper]{article} 
\usepackage{aaai25}  
\usepackage{times}  
\usepackage{helvet}  
\usepackage{courier}  
\usepackage[hyphens]{url}  
\usepackage{graphicx} 
\def\UrlFont{\rm}  
\usepackage{natbib}  
\usepackage{caption} 
\usepackage{algorithm}
\usepackage{algorithmic}

\usepackage{newfloat}
\usepackage{listings}
\DeclareCaptionStyle{ruled}{labelfont=normalfont,labelsep=colon,strut=off} 
\floatstyle{ruled}
\newfloat{listing}{tb}{lst}{}
\floatname{listing}{Listing}
\newcounter{definition}
\newtheorem{Definition}[definition]{Definition}

\usepackage{amsmath}
\usepackage{amssymb}
\usepackage{pifont}
\usepackage{caption}
\usepackage{subcaption}
\usepackage{gensymb}
\usepackage{dsfont}
\usepackage{subcaption}
\usepackage{etoolbox}
\usepackage{booktabs}

\usepackage{algorithm,algorithmic}

\title{Anomalous Agreement: How to find the Ideal Number of Anomaly Classes in Correlated, Multivariate Time Series Data}
\author {
    Ferdinand Rewicki\textsuperscript{\rm 1,\rm 2},
    Joachim Denzler\textsuperscript{\rm 2}
    Julia Niebling\textsuperscript{\rm 1},
}
\affiliations {
    \textsuperscript{\rm 1}Institute of Data Science, German Aerospace Center, 07745 Jena, Germany \\
    \textsuperscript{\rm 2}Computer Vision Group, Friedrich-Schiller University, 07743 Jena, Germany \\
    ferdinand.rewicki@dlr.de, joachim.denzler@uni-jena.de, julia.niebling@dlr.de
}

\nocopyright

\begin{document}
\maketitle

\begin{abstract}
Detecting and classifying abnormal system states is critical for condition monitoring, but supervised methods often fall short due to the rarity of anomalies and the lack of labeled data. Therefore, clustering is often used to group similar abnormal behavior. However, evaluating cluster quality without ground truth is challenging, as existing measures such as the Silhouette Score (SSC) only evaluate the cohesion and separation of clusters and ignore possible prior knowledge about the data. To address this challenge, we introduce the Synchronized Anomaly Agreement Index (SAAI), which exploits the synchronicity of anomalies across multivariate time series to assess cluster quality. We demonstrate the effectiveness of SAAI by showing that maximizing SAAI improves accuracy on the task of finding the true number of anomaly classes $K$ in correlated time series by $0.23$ compared to SSC and by $0.32$ compared to X-Means. We also show that clusters obtained by maximizing SAAI are easier to interpret compared to SSC.
\end{abstract}

%
 \begin{links}
     \link{Code}{https://gitlab.com/dlr-dw/saai}
 \end{links}

\section{Introduction}
Detecting and classifying abnormal system states is crucial for effective monitoring and control of complex systems. Unfortunately, supervised classification approaches fall short because anomalies are by definition rare, and especially for real-world applications, no or very limited labeled data is available. Therefore, clustering is used to derive groups of similar anomalous behavior from unlabeled data \cite{Sohn2023, Rewicki2024}. Assessing the quality of a given solution obtained by applying clustering algorithms such as K-means \cite{MacQueen1967} to the anomalous subsequences or inferred features is challenging for a number of reasons: (a) no ground truth is available to determine the quality of a solution, (b) the true number of clusters in the data is usually unknown, (c) the solution is highly dependent on the chosen embedding in a feature space \cite{Rewicki2024,Raihan2023}. Furthermore, classical unsupervised cluster quality measures such as the Silhouette Score (SSC) \cite{Rousseeuw1987} evaluate the cohesion within and the separation between clusters but do not incorporate any prior knowledge about the data. To address this challenge in the case of multivariate time series consisting of sufficiently similar signals, we investigate the following research question: How can we exploit the similarity between signals when clustering anomalies found in these variables? 
The SAAI is based on the principle, that anomalies found simultaneously (i.e., synchronously) in several similar variables within a multivariate time series should belong to the same class.
In this work, we deliver evidence on the effectiveness of this measure and show, that maximising SAAI is superior compared to maximizing SSC and the X-Means algorithm \cite{Pelleg2000}, a variant of K-Means that determines the ideal value for K. Our contributions are:
\begin{enumerate}
    \item We derive the SAAI, an internal measure of the quality of anomaly clusters.
    \item We justify the effectiveness of SAAI by showing that SAAI outperforms SSC and X-Means on the task of finding the true number of classes $K$. We show that the results obtained by using SAAI are highly correlated with those obtained by using the Adjusted Rand Index (ARI) \cite{Hubert1985} and the Fowlkes Mallows Index (FMI) \cite{FowlkesMallows1983}, two external cluster quality measures that require ground truth labels.
    \item We show that the clusterings obtained from maximizing SAAI are easier to interpret compared to SSC.
\end{enumerate}

The rest of the paper is organized as follows: We start with discussing related work in Section~\ref{sec:relatedwork}. In Section~\ref{sec:method} we derive the SAAI using an illustrative example and introduce the synthetic dataset. In Section~\ref{sec:results} we present the experimental setup
and results, which we discuss in Section~\ref{sec:discussion}. Finally, in Section~\ref{sec:conclusion} we conclude and give an outlook on future work.

\section{Related Work}
\label{sec:relatedwork}
The Silhouette Score, introduced in \cite{Rousseeuw1987}, is the standard measure for evaluating clustering results and quantifies both, cohesion and separation within clusters. 
It is calculated by averaging the silhouette coefficients $SSC_{C_j}$ for each cluster $C_j$, defined as
\begin{equation} 
\label{eq:ssc} 
  SSC_{C_j}(C_j) = \frac{1}{|C_j|} \sum_{\mathcal{S} \in C_j} \frac{idist(\mathcal{S}) - wdist(\mathcal{S})}{\max(wdist(\mathcal{S}), idist(\mathcal{S}))} \text{.}
\end{equation}
The measure $wdist(\mathcal{S})$ is the average distance of the object $\mathcal{S} \in C_j$ to all other elements within its own cluster $C_j$ (within-cluster distance), while $idist(\mathcal{S})$ is the smallest average distance to elements in another cluster $C_i \neq C_j$ (inter-cluster distance).
The SSC ranges from $-1$ to $1$, where $1$ indicates well-separated clusters, $0$ indicates overlapping clusters, and $-1$ indicates misclassification of objects.
As SSC does only use information obtained from the clustering process, it is referred to as an \textit{internal} measure.

The Fowlkes-Mallows Index \cite{FowlkesMallows1983} is an external measure of the similarity between two clusterings $C_i$ and $C_j$, i.e. two partitions of a finite set of objects. An external measure uses information obtained from outside the clustering process, e.g. ground truth class labels. The FMI is the geometric mean of the product of Precision and Recall, ensuring a balanced evaluation of the two quantities. It also defines a scalar product on the space of pairs of data points \cite{BenHur2002}.  We use FMI to measure the similarity between a clustering and the ground truth class assignments. 
The FMI is defined as
\begin{equation}
    FMI = \sqrt{\frac{TP}{TP+FP} \frac{TP}{TP+FN}} \text{ ,}
\end{equation}
where TP is the number of true positives, FP is the number of false positives and FN is the number of false negatives. 
The FMI ranges from $0$ to $1$, where $FMI = 0$ indicates absolute disagreement and $FMI = 1$ perfect similarity of $\mathcal{C}_i$ and $\mathcal{C}_j$. 
In the context of comparing a predicted to a true clustering, a true positive is a pair of points that belongs to the same cluster in both, the predicted and the true solution. 
A false positive is a pair of points that belongs to the same cluster in the predicted, but to different clusters in the true clustering. 
False negative and true negative are defined accordingly.  

The Rand Index (RI) \cite{Rand1971} is another external measure of the similarity between two clusterings and represents the ratio of correct decisions to all decisions. The RI is defined as
\begin{equation}
    RI = \frac{TP + TN}{TP+FP+FN+TN} \text{ ,}
\end{equation}
where TP and TN are the number of true positives and true negatives and FP and FN are the number of false positives and false negatives. 
\citeauthor{Hubert1985} proposed the Adjusted Rand Index (ARI) in \cite{Hubert1985}, which corrects the original RI by accounting for the expected similarity of random cluster assignments, making the measure robust against chance agreement. This correction is achieved by subtracting the expected value of the RI and dividing by its maximum minus the expectation \cite{Hubert1985}. ARI ranges from $-1$ to $1$, where an index of $1$ represents perfect agreement, $-1$ perfect disagreement and $0$ the expected agreement by chance.

A popular heuristic for finding the ideal number of clusters is the Elbow-method \cite{Kodinariya2013,Bholowalia2014,Lopez2018}. This is a visual method that plots the sum of squared errors of objects to their assigned cluster centers against increasing number of clusters. The value where the curve forms an elbow is selected as the ideal value. However, this method is highly subjective and there is no guarantee that an elbow point can be identified.\cite{Schubert2023} Therefore, various quantitative methods for selecting the ideal number of clusters have been proposed. 

\cite{Dinh2019} and \cite{ShahapureNicholas2020} propose methods to find the ideal number of clusters by maximizing the SSC. \cite{Dinh2019} test their approach on categorical data and use a Lin-similarity based measure for categorical data proposed in \cite{Nguyen2019}. \cite{ShahapureNicholas2020} evaluate their approach on continuous non-time series data. Both works find that maximizing the SSC yields the correct number of classes in their experiments.

\citeauthor{Raihan2023} proposes a method for finding the ideal number of clusters for time series datasets using a symbolic pattern forest algorithm in \cite{Raihan2023}. The experiments show, that SSC fails on finding the correct number of clusters when working with raw time series. Although this observation is presented without an explanation, it underpins our observation of the shortcomings of SSC, esp. in combination with raw time series.

\citeauthor{Pelleg2000} proposed X-Means in \cite{Pelleg2000}, a variant of K-Means that does not need the number of clusters be selected in advance but finds it by optimizing either the Bayesian Information Criterion (BIC) or the Akaike Information Criterion (AIC). To compare our proposed method to X-Means, we use a time series compatible version of X-Means, in which we replaced the euclidean distance with the elastic distance measure Merge-Split-Merge (MSM) \cite{Stefan2013} to compare time series of unequal length.

While maximizing SSC has been shown to work well for non-time series data to find the ideal number of clusters, its usefulness for clustering raw time series is questionable. This leaves a gap for different approaches, which we contribute to fill with this work.

In our earlier work \cite{Rewicki2024}, we proposed a methodology for deriving anomaly types from unlabeled time series data and introduced the SAAI in an informative way. This work is intended to provide evidence of its usefulness and performance.

\section{Methodology}
\label{sec:method}
In the following section we present the methodology of this study. We start with  deriving the SAAI and give an illustrative example. Afterwards we introduce the synthetic dataset, which we use in our experiments to justify the proposed measure.

\subsection{Synchronized Anomaly Agreement Index (SAAI)}
\label{sec:method:saai}

To evaluate the quality of clustering results, we introduce the \textbf{Synchronized Anomaly Agreement Index (SAAI)}. The rationale behind this measure is to use prior knowledge about the signals of a multivariate time series. Assuming sufficiently correlated signals, synchronized, i.e. temporally aligned, anomalies in different channels should be assigned to the same cluster, as they are likely to represent the same anomaly type. 

We begin with the basic definitions:
\begin{Definition}
The regular \textbf{time series} $\mathcal{T}$ of length $N \in \mathbb{N}$ is defined as the set of pairs {${\mathcal{T} = \lbrace{(t_n, \mathbf{x_n}) | t_n \leq t_{n+1}, 0 \leq n \leq N-1, t_{n+1} - t_n = c \rbrace}}$} with $\mathbf{x_n} \in \mathbb{R}^D$ being the data points with $D$ behavioral attributes and $t_n \in \mathbb{N}, n \leq N$ being the equidistant timestamps with distance $c$ to which the data refer. 
For $D=1$, $\mathcal{T}$ is called univariate, and for $D>1$, $\mathcal T$ is called multivariate.
\end{Definition}

Since we discuss the matter of clustering anomalous subsequences of a time series, we define a subsequence as a connected subset of $\mathcal{T}$: 
\begin{Definition}
The \textbf{subsequence} $\mathcal{S}_{a,b} \subseteq \mathcal{T}$ of the time series $\mathcal{T}$, with length $L = b-a+1 > 0$ is given by {$\mathcal{S}_{a,b} := \lbrace{ (t_n, \mathbf{x_n}) | 0 \leq a \leq n \leq b \leq N-1 \rbrace }$}. 
For multivariate time series $\mathcal{T}$, $\mathcal{S}^{(i)}_{a,b}$ with $i \in \mathbb{N}$ refers to the subsequence $\mathcal{S}_{a,b}$ in dimension $1 \leq i \leq D$
For brevity, we often omit the indices and refer to arbitrary subsequences as $\mathcal{S}$.
\end{Definition}

We continue with the definition of anomalies and synchronized anomalies:
\begin{Definition}
Given the time series $\mathcal{T}$ with $D > 1$ and a subsequence $\mathcal{S}^{(i)}_{a,b}$, the set $A$ of \textbf{univariate, anomalous subsequences} is given as
\begin{equation}
    A := \{\mathcal{S}^{(i)}_{a,b} | i, a, b \in \mathbb{N}, i \leq D, a < b, s(\mathcal{S}^{(i)}_{a,b}) \geq \theta_s\} \text{   ,}
\end{equation}
with $s(\cdot): \{\mathcal{S} | \mathcal{S} \subseteq \mathcal{T}\} \rightarrow [0,1]$ being an anomaly score function and $\theta_s \in [0,1]$ being the threshold to classify a subsequence $\mathcal{S}^{(i)}_{a,b}$ as anomalous.
\end{Definition}

\begin{Definition}
\label{def:syncanom}
Let $\mathcal{S}^{(i)}_{a_i,b_i}, \mathcal{S}^{(j)}_{a_j,b_j}$ with $i < j$ be two univariate subsequences in two different dimensions of the multivariate time series $\mathcal{T}$. We say $\mathcal{S}^{(i)}_{a_i,b_i}$ and $\mathcal{S}^{(j)}_{a_j,b_j}$ are \textbf{synchronized}, if they overlap by more than a threshold $\theta$:
\begin{equation}
\label{eq:syncanom}
    \omega(a_i,b_i,a_j,b_j) := \frac{min(b_i,b_j)-max(a_i,a_j)}{max(b_i,b_j)-min(a_i,a_j)} \geq \theta
\end{equation}
with $\theta \in [0,1]$.

The set of all synchronized, univariate, anomalous subsequences $A_S$ is given as
\begin{equation}
\begin{aligned}
    A_S := \{& (\mathcal{S}^{(i)}_{a_i,b_i}, \mathcal{S}^{(j)}_{a_j,b_j}) | \mathcal{S}^{(i)}_{a_i,b_i}, \mathcal{S}^{(j)}_{a_j,b_j} \in A, \\
        & i < j, \omega(a_i,b_i,a_j,b_j) \geq \theta \}
    \end{aligned}
\end{equation}
\end{Definition}


\begin{Definition}
Let $A_S$ be the set of all synchronized, univariate, anomalous subsequences of time series $\mathcal{T}$. The subset $A^*_S \subseteq A_S$ of \textbf{synchronized anomalies that agree on their cluster} is given as 
\begin{equation}
    A^*_S = \{ (\mathcal{S}^{(i)}, \mathcal{S}^{(j)}) | (\mathcal{S}^{(i)}, \mathcal{S}^{(j)}) \in A_S, c(\mathcal{S}^{(i)}) = c(\mathcal{S}^{(j)}) \} \text{   ,}
\end{equation}
with $c(\mathcal{S})$ denoting the cluster of subsequence $\mathcal{S}$
\end{Definition}

\begin{Definition}
\label{def:saai}
Given the set of synchronized, univariate anomalous subsequences $A_S$ and the set of synchronized, univariate anomalous subsequences in the same cluster $A^*_S$, the number of clusters $K$ and the number of pseudo-clusters containing only a single element $n_\mathds{1}$, the \textbf{Synchronized Anomaly Agreement Index (SAAI)} is defined as:
\begin{equation}
\label{eq:saai}
    SAAI := \lambda \frac{|A^*_S|}{|A_S|} + (1-\lambda)\frac{K-1-n_\mathds{1}}{K} \text{ , } \lambda \in [0,1] \text{  . } 
\end{equation}
Here, the first term $\frac{|A^*_S|}{|A_S|}$ evaluates the ratio of synchronized anomalies in the same cluster to all synchronized anomalies.
The second term serves as a regularization to account for small cluster sizes ($\frac{1}{K}$) and clusters containing only a single anomaly ($\frac{n_\mathds{1}}{K}$). It also ensures that the SAAI is in $[0,1]$. The parameter $\lambda$ allows to adjust the influence of the regularizer on the main term. 
\end{Definition}
The algorithm to calculate the SAAI, its complexity analysis and further information on selecting $\lambda$ can be found in the Appendix.

\begin{figure}[ht]
    \centering
    \begin{subfigure}{0.49\columnwidth}
        \centering
        \includegraphics[width=\linewidth]{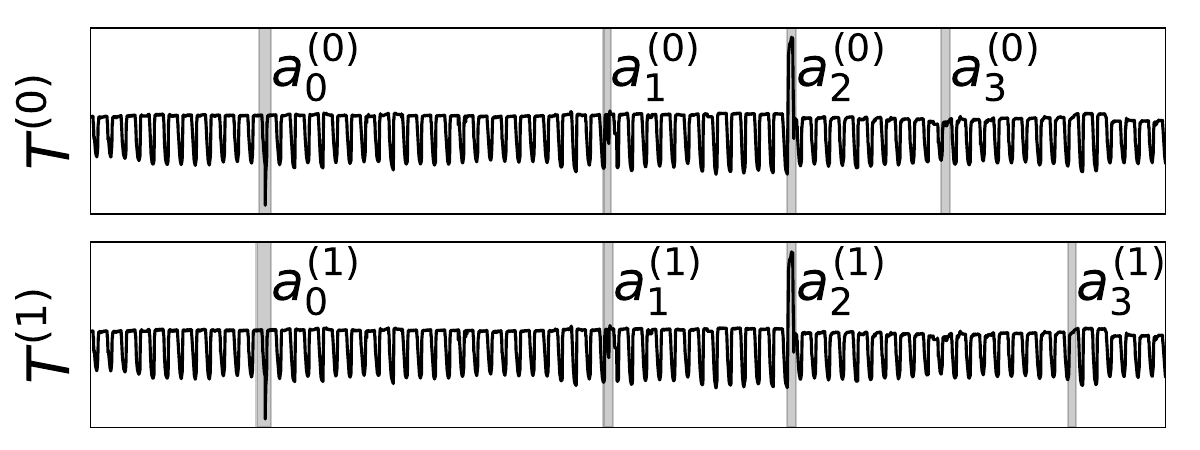}
        \caption{Detected Anomalies}
        \label{fig:saai:ex1}
    \end{subfigure}
    \hfill
    \begin{subfigure}{0.49\columnwidth}
        \centering
        \includegraphics[width=\linewidth]{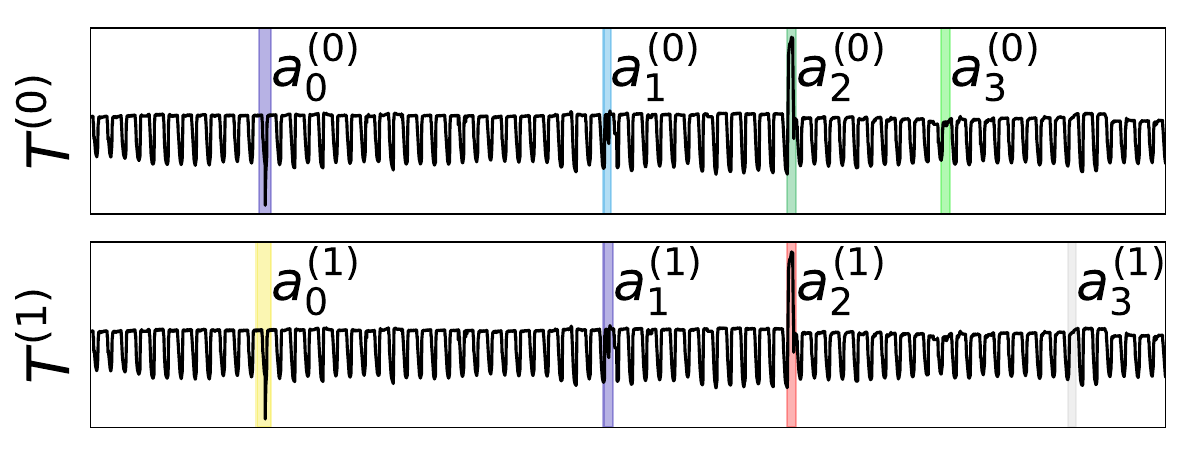}
        \caption{$SAAI = 0$}
        \label{fig:saai:ex6}
    \end{subfigure}
    \begin{subfigure}{0.49\columnwidth}
        \centering
        \includegraphics[width=\linewidth]{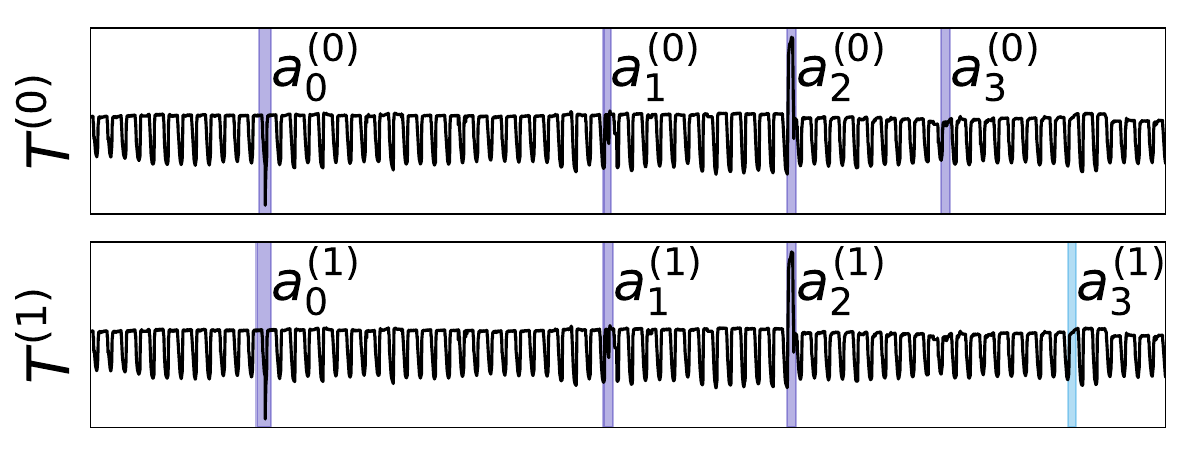}
        \caption{$SAAI = 0.5$}
        \label{fig:saai:ex2}
    \end{subfigure}
    \hfill
    \begin{subfigure}{0.49\columnwidth}
        \centering
        \includegraphics[width=\linewidth]{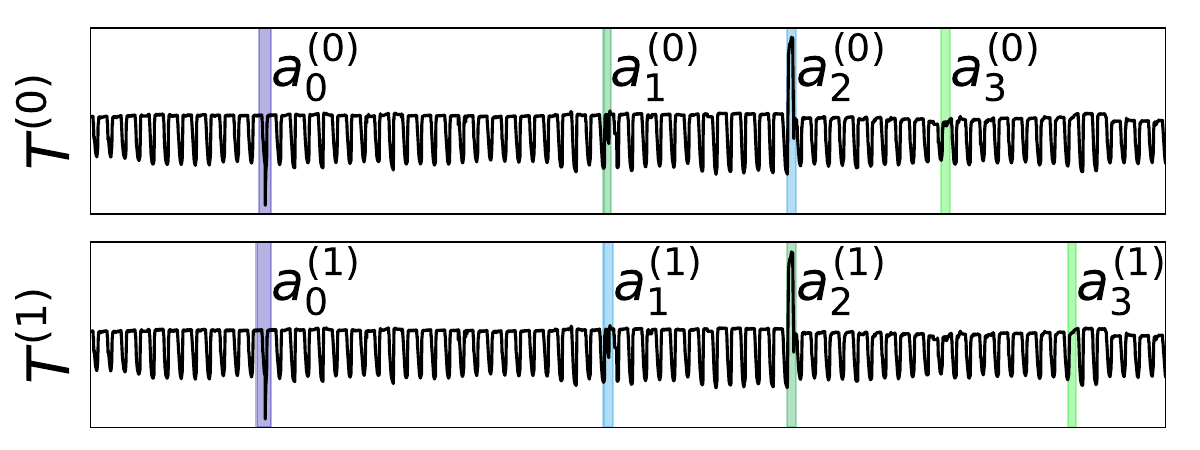}
        \caption{$SAAI = 0.541\bar{6}$}
        \label{fig:saai:ex5}
    \end{subfigure}
    \begin{subfigure}{0.49\columnwidth}
        \centering
        \includegraphics[width=\linewidth]{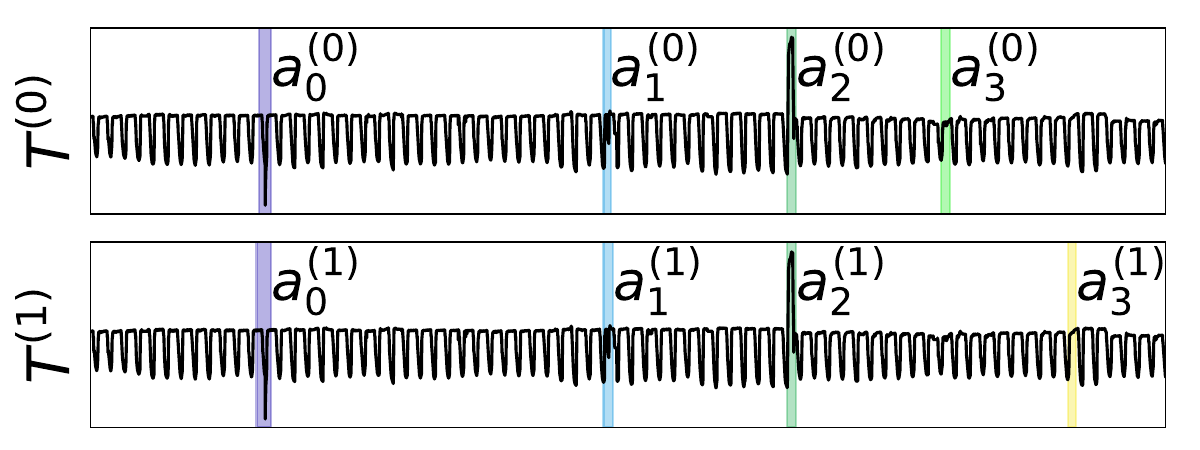}
        \caption{$SAAI = 0.7$}
        \label{fig:saai:ex4}
    \end{subfigure}
    \hfill
    \begin{subfigure}{0.49\columnwidth}
        \centering
        \includegraphics[width=\linewidth]{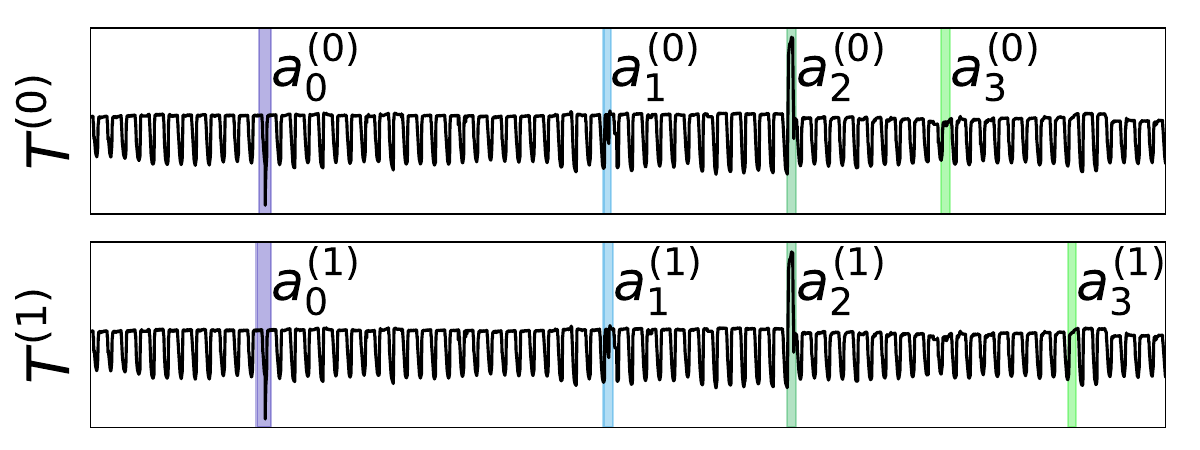}
        \caption{$SAAI = 0.875$}
        \label{fig:saai:ex3}
    \end{subfigure}
    
    \caption{(a) the detected anomalies $a^{(i)}_j$ and (b) - (f) different clustering solutions with increasing quality. Cluster assignment is coded by color. (b) Worst case: all but one cluster contain a single element, (c) all but one anomaly assigned to the same cluster, (d) synchronized anomalies not in the same cluster, (e) synchronized anomalies in separate clusters, pseudo-clusters exist, (f) best case: synchronized anomalies in separate clusters, no pseudo-cluster.}
    \label{fig:saai:examples}
\end{figure}

\paragraph{Example}
Figure \ref{fig:saai:examples} illustrates how the SAAI is calculated for different clusterings. In the following example we use $\lambda=0.5$ and $\theta=0.5$. A high resolution version of Figure~\ref{fig:saai:examples} can be found in the appendix.
Figure \ref{fig:saai:ex1} shows the two time series $\mathcal{T} = \lbrace{T^{(1)}, T^{(2)}\rbrace}$  and the detected anomalies $A = \lbrace{ a^{(i)}_j | i \in \lbrace{ 0,1 \rbrace}, j \in \lbrace{ 0, 1, 2, 3 \rbrace} \rbrace}$. 
The SAAI is calculated over the set of synchronized anomalies: $A_S = \lbrace{(a^{(0)}_0, a^{(1)}_0), (a^{(0)}_1, a^{(1)}_1), (a^{(0)}_2, a^{(1)}_2) \rbrace}$, $|A_S|=3$.
The anomalies $a^{(0)}_3$ and $a^{(1)}_3$ are not synchronized and hence $((a^{(0)}_3, a^{(1)}_3)) \notin A_S$.

Figure \ref{fig:saai:ex6} shows the worst-case solution with the lowest possible SAAI value of $0$, where all but two unsynchronized anomalies are assigned to separate clusters. 
We are aware that this is an extreme edge case, although it is perfectly valid.

Figure \ref{fig:saai:ex2} shows another extreme case, where all but one unsynchronized anomaly are assigned to the same cluster. 
Here, in particular, the synchronized anomalies are assigned to the same cluster, which is the goal of the main term in Equation (\ref{eq:saai}), but the information contained in this clustering is low, which is regularized by the first part of the penalty term $\frac{1}{k}$. 
The SAAI of this solution is $0.5$.

The solution shown in Figure \ref{fig:saai:ex5} assigned the synchronized anomalies $(a^{(0)}_0, a^{(1)}_0)$ to the same cluster, the remaining synchronized anomalies are assigned to different clusters and no pseudo-cluster with only a single element is contained. 
The SAAI value of this solution is slightly higher with $0.541\bar{6}$ as in \ref{fig:saai:ex2}.

Figure \ref{fig:saai:ex4} shows the near perfect solution where all synchronized anomalies are assigned to the same cluster, but the different anomaly types are assigned to different clusters. 
However, two pseudo-clusters are included in this solution, resulting in an SAAI value of $0.7$.

The best case solution for this example is shown in Figure \ref{fig:saai:ex3}, which is similar to the one shown in Figure \ref{fig:saai:ex4}, but no pseudo clusters are included in this solution, giving a SAAI value of $0.875$. 
The maximal value of $SAAI=1$ could be reached if we set $\lambda=0$.

\subsection{Synthetic Dataset}
\label{sec:method:synthdata}
For the experiments in Section~\ref{sec:res:exp1}, we created synthetic time series similar to temperature measurements from the EDEN ISS \cite{Zabel2017} Illumination Control System (ICS) in the EDEN ISS 2020 telemetry dataset. EDEN ISS was a research greenhouse for the study of Controlled Environment Agriculture (CEA) techniques and plant growth in (semi)-closed environments, operating between 2018 and 2021 in Antarctica, near the German Neumayer III polar station.

\begin{figure}[ht]
    \centering
    \begin{subfigure}{.98\columnwidth}
        \centering
        \includegraphics[width=\linewidth]{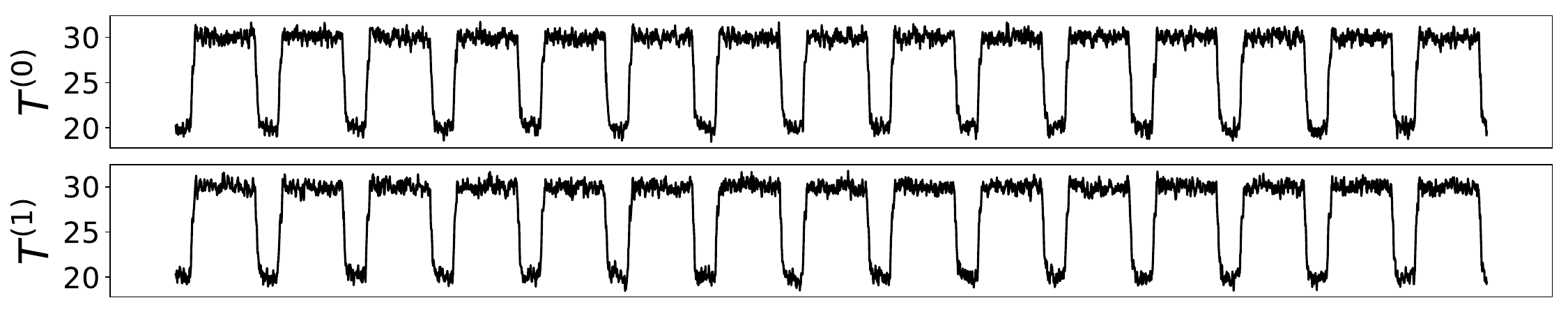}
        \caption{}
        \label{fig:data:raw:noise_added}
    \end{subfigure}
    \begin{subfigure}{.98\columnwidth}
        \centering
        \includegraphics[width=\linewidth]{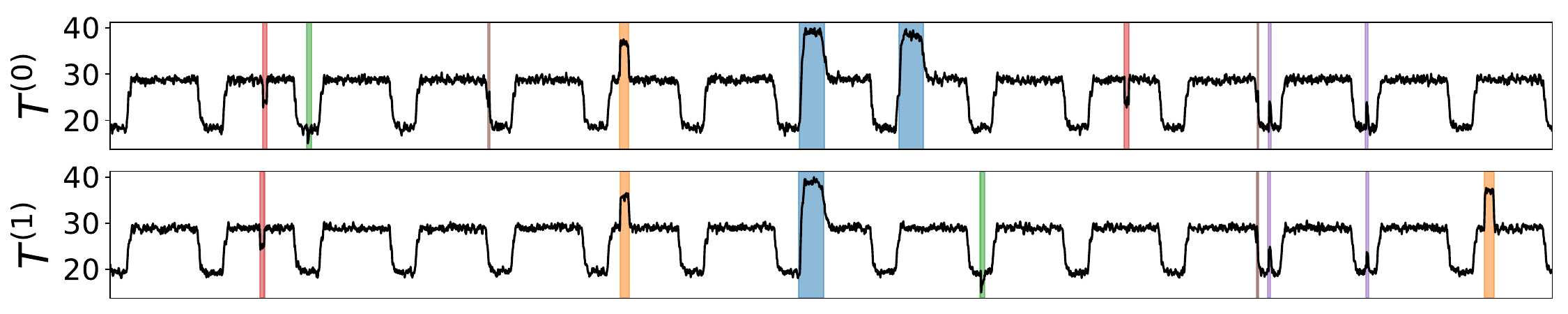}
        \caption{}
        \label{fig:data:raw:with_anomalies}
    \end{subfigure}
    \caption{(a): The basic synthetic ICS signal with simulated sensor noise, (b) The synthetic ICS signal with injected anomalies and $r_{sync} = 0.8$.}
    \label{fig:data:raw}
\end{figure}

The synthetic time series consists of a periodic signal mimicking the regular ICS temperature signal following the illumination pattern of EDEN ISS with $6h$ night phase at $20 \degree C$, $1h$ warm-up during the simulated sunrise with a small overshoot above the desired daytime temperature of $30 \degree C$. 
The warm-up is followed by $16h$ of daytime at $30 \degree C$ and finally $1h$ of cool-down. 
To simulate sensor noise, we add red noise with zero mean, $0.5$ standard deviation, and a correlation coefficient of $0.5$ to the signal. 
The basic noisy signal and an example with injected anomalies are shown in Figure~\ref{fig:data:raw}.

\begin{figure}[ht]
    \centering
    \captionsetup{justification=centering}
    \begin{subfigure}[t]{.15\columnwidth}
        \includegraphics[width=\textwidth]{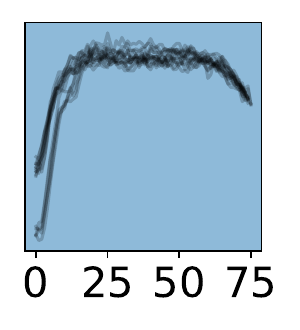}    
        \caption{Long Peak}
        \label{feg:dataset:class0}
    \end{subfigure}
    \begin{subfigure}[t]{.15\columnwidth}
        \includegraphics[width=\textwidth]{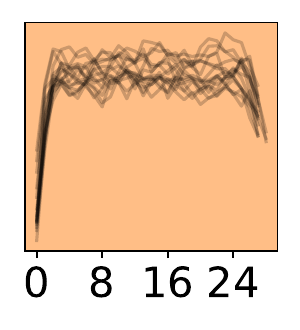}    
        \caption{Short Peak}
        \label{feg:dataset:class1}
    \end{subfigure}
    \begin{subfigure}[t]{.15\columnwidth}
        \includegraphics[width=\textwidth]{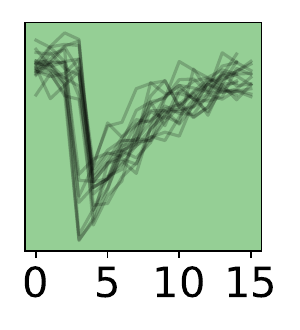}    
        \caption{Night Drop}
        \label{feg:dataset:class2}
    \end{subfigure}
    \begin{subfigure}[t]{.15\columnwidth}
        \includegraphics[width=\textwidth]{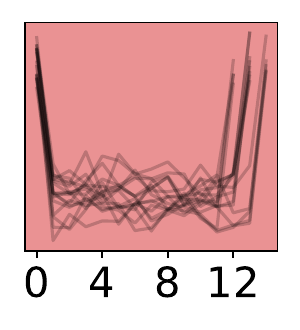}    
        \caption{Day Drop}
        \label{feg:dataset:class3}
    \end{subfigure}
    \begin{subfigure}[t]{.15\columnwidth}
        \includegraphics[width=\textwidth]{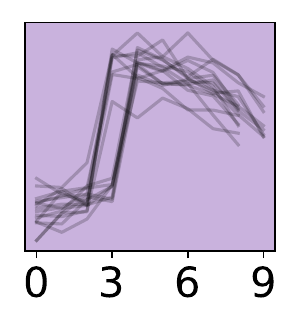}    
        \caption{Night Peak}
        \label{feg:dataset:class4}
    \end{subfigure}
    \begin{subfigure}[t]{.15\columnwidth}
        \includegraphics[width=\textwidth]{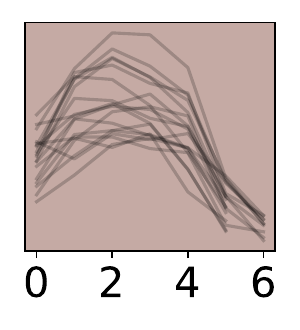}    
        \caption{Cool- down Peak}
        \label{feg:dataset:class5}
    \end{subfigure}
    \caption{The six anomaly types that are injected into the base signal.}
    \label{fig:data:sequences}
\end{figure}

To validate the SAAI, we inject 6 different anomaly types into the raw signal, which are shown in Figure \ref{fig:data:sequences} and described in Table \ref{tab:dataset:anomalies}. 
These anomaly types have also been observed in the real ICS temperature signals as described in  \cite{Rewicki2024} and are considered to be distinct types of anomalous behavior. 
Each injected anomaly is subject to randomness with respect to start time, duration, and intensity. 
Another degree of freedom is the ratio of synchronized to unsynchronized anomalies. 
This ratio is steered via a parameter $r_{sync}$. Setting $r_{sync} = 1$ yields a multivariate time series with synchronized anomalies only, while setting $r_{sync} = 0$ yields no synchronized anomaly at all.

\begin{table}[]
    \centering
    \fontsize{9pt}{11pt}\selectfont  
    \begin{tabular}{@{}llrr@{}}
    \toprule
    \textbf{Name}  & \textbf{Start Time} & \textbf{Dur. (min)} & \textbf{Intensity} \\ \midrule
    Long Day Peak  & 04:00 - 06:20 & 240 - 245 & $10\degree C - 11\degree C$ \\
    Short Day Peak & 07:00 - 08:20 & 120 - 125 & $8\degree C - 9\degree C$   \\
    Night Drop     & 01:00 - 01:40 & 10 - 11 & $-5\degree C - -4\degree C$  \\
    Day Drop       & 13:00 - 15:50 & 60 - 65 & $-5\degree C - -4\degree C$\\
    Night Peak     & 01:00 - 01:40 & 10 - 11 & $5\degree C - 6\degree C$   \\
    Cooldown Peak  & 22:00 - 22:30 & 20 - 21 & $5\degree C - 6\degree C$   \\ \bottomrule
    \end{tabular}
    \caption{Parametrization of the six anomaly types, we inject into the synthetic ICS signal.}
    \label{tab:dataset:anomalies}
\end{table}

\section{Experiments and Results}
\label{sec:results}
\subsection{Experimental Setup}
We implemented our experiments using Python (version 3.11). 
For Clustering we used the K-Means implementation in the TSLearn package \cite{TSLearn}, which is compatible with subsequences of unequal length. 
For the same reason, we implemented the X-Means algorithm to use K-Means with an elastic distance measure. 
We found that we get better results when using Merge-Split-merge (MSM) \cite{Stefan2013} compared to using Dynamic Time Warping (DTW) \cite{Vintsyuk1968,Sakoe1978,Berndt1994} so we compare to the X-Means version with MSM as distance measure. 
We also noticed a major difference in our results depending on the SSC implementation. 
We compared that in the TSLearn package to that in scikit-learn \cite{scikit-learn} with the DTW implementation from the aeon-toolkit \cite{aeon24jmlr} and found that the SSC in scikit-learn yields much better results, even though it is not compatible with unequal length subsequences. Therefore we padded the subsequences with zero to make them equal length. A comparison of these two SSC variants can be found in the Appendix.
All experiments were run on an Intel Xeon Platinum 8260 CPU with 5GB of allocated memory

\subsection{Synthetic Greenhouse Temperature Data}
\label{sec:res:exp1}
To evaluate SAAI on finding the ideal number of classes $K$ within multivariate time series containing different types of anomalies, we perform the following experiments: We generate a large number of multivariate time series of the synthetic ICS temperature measurements and vary different parameters, namely the number ($K$) and type of injected anomaly classes, the dimension ($D$) of the multivariate time series, and the ratio of synchronized to unsynchronized anomalies $r_{sync}$. We then cluster the anomalous sequences using $K$-Means clustering with $(DTW)$ as the distance measure. To remove high-frequent noise from the sequences, we apply moving average smoothing with a window size of~$5$. For each parameter we change, we generate 50 multivariate time series and cluster the anomalous subsequences of each time series with $2 \leq k < 20$, where $k$ is the number of clusters for K-means. We measure how often the correct value $K$ was found by maximizing the internal metrics SAAI and SSC. In addition, we compute the external metrics SAAI and SAAI and use them to find the true number of classes, again by maximizing their respective values. It is fair to complain that with access to the ground truth labels, finding the true number of classes $K$ by maximizing an external metric is pointless. However, we do this for the sake of analyzing the correlation of the internal metrics SSC and SAAI with the external metrics ARI and FMI. As another competitor, we use the X-Means Algorithm \cite{Pelleg2000} to determine the ideal value for $K$.

\begin{figure}[ht]
    \centering
    \begin{subfigure}{.32\columnwidth}
        \centering
        \includegraphics[width=\columnwidth]{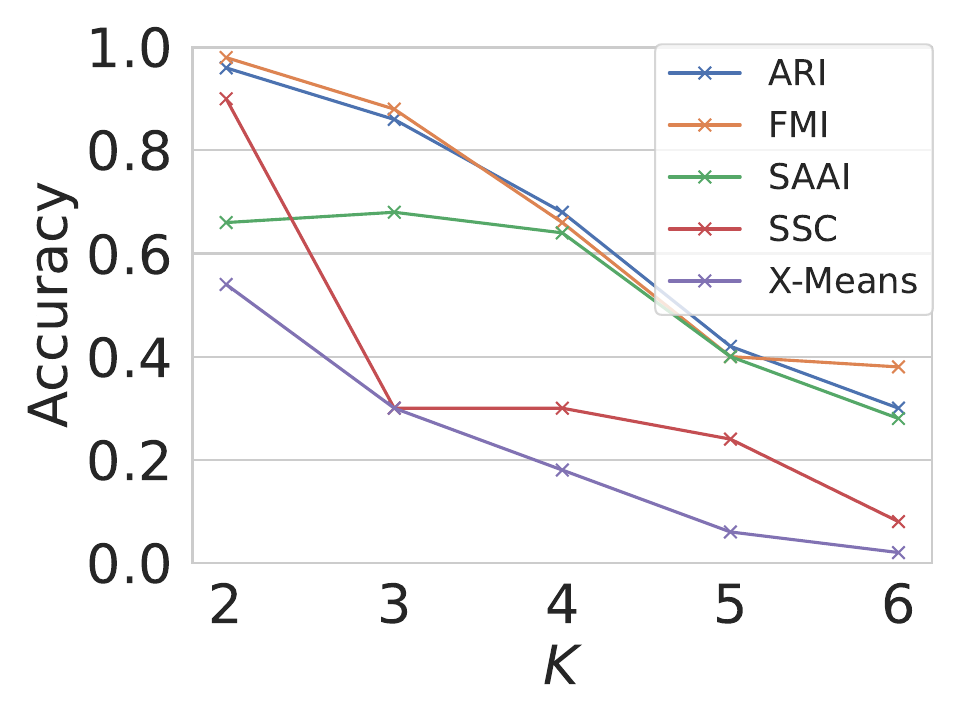}
        \caption{Increasing $K$}
        \label{fig:res:synth:inc_k}
    \end{subfigure}
    \hfill
    \begin{subfigure}{.32\columnwidth}
        \centering
        \includegraphics[width=\columnwidth]{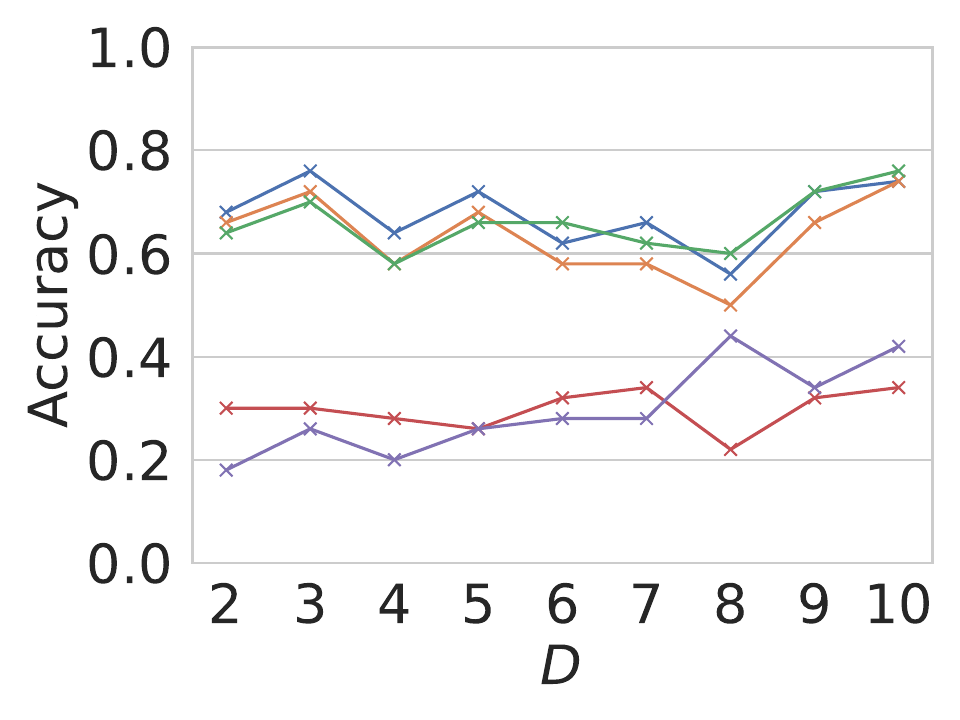}
        \caption{Increasing $d$}
        \label{fig:res:synth:inc_d}
    \end{subfigure}
    \hfill
    \begin{subfigure}{.32\columnwidth}
        \centering
        \includegraphics[width=\columnwidth]{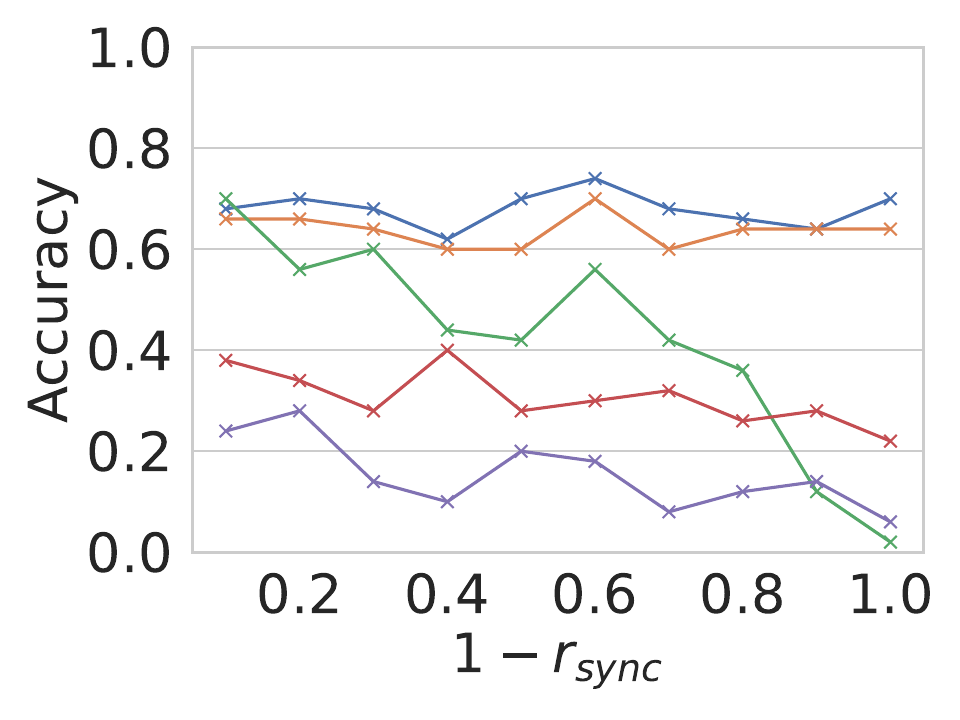}
        \caption{Decreasing $r_{sync}$}
        \label{fig:res:synth:inc_rsync}
    \end{subfigure}
    \caption{Results of the experiments on synthetic ICS data as described in Section \ref{sec:res:exp1}. Except for $K=2$ and $r_{sync} < 0.2$, maximizing SAAI is superior to maximizing SSC. X-Means beats SAAI only for $r_{sync} < 0.2$.}
    \label{fig:res:synth:lines}
\end{figure}

\paragraph{Increasing $K$}
In the first experiment we fix $D=2$, choose $r_{sync} \in [0.5, 1]$ and increase the number of classes from $K = 2$ to $K = 6$. We run the experiment 50 times for each value of $K$ and select a new value for $r_{sync}$ as well as $K$ new classes on each run uniformly at random. Figure \ref{fig:res:synth:inc_k} shows the accuracy in finding the true number of classes for increase $K$. 
Except for $K=2$, SAAI shows superior performance compared to SSC. For $K>3$, SAAI is almost as good as using the external metrics ARI and FMI. X-Means shows the worst results among the compared methods. All methods show decreasing accuracy as $K$ increases.

\paragraph{Increasing $D$}
Now, we increase the dimension of the multivariate time series from $D=2$ to $D=10$ while fixing $K=4$ and choosing $r_{sync} \in [0.5, 1]$ . We run the experiment 50 times for each value for $d$ and select a new value for $r_{sync}$ as well as new $K=4$ classes on each run uniformly at random. Figure \ref{fig:res:synth:inc_d}  shows the accuracy in finding the true number of classes $K$ for increasing dimension $D$. 
Again, SAAI shows superior performance in finding the true value $K$ compared to SSC and X-Means.
Compared to ARI and FMI, SAAI is almost on par with the external metrics for $D < 6$ and even slightly better for $D \geq 6$.
Despite the minor variability in accuracies within the results for one method, we also see that the accuracy of finding the true value $K$ is almost independent from the dimension $D$ of the multivariate time series.

\paragraph{Decreasing $r_{sync}$}
In the third experiment, we fix $K=4$ and $D=2$ and decrease $r_{sync}$ from $r_{sync} = 1$ to $r_{sync} = 0$ in steps of $0.1$.  We run the experiment 50 times for each value of $r_{sync}$ and select $K$ new classes on each run uniformly at random. 
Figure \ref{fig:res:synth:inc_rsync} shows the accuracy in finding the true number of classes $K$.  
As in the previous experiments, SAAI proves to be superior to SSC and X-Means in determining the true value $K$. Compared to ARI and FMI, the accuracy for SAAI is on par for $1-r_{sync} < 0.2$. For $0.2 \leq 1-r_{sync} \leq 0.8$, the accuracy for SAAI shows an decreasing trend as expected, but is still higher than for SSC and X-Means. For $1-r_{sync} > 0.8$ the accuracy for SAAI falls below that of SSC and X-means
The correlation of SAAI and $r_{sync}$ is expected, since $r_{sync}$ determines the proportion of synchronized anomalies in the time series. 

\begin{figure}[ht]
    \centering
    \includegraphics[width=.9\columnwidth]{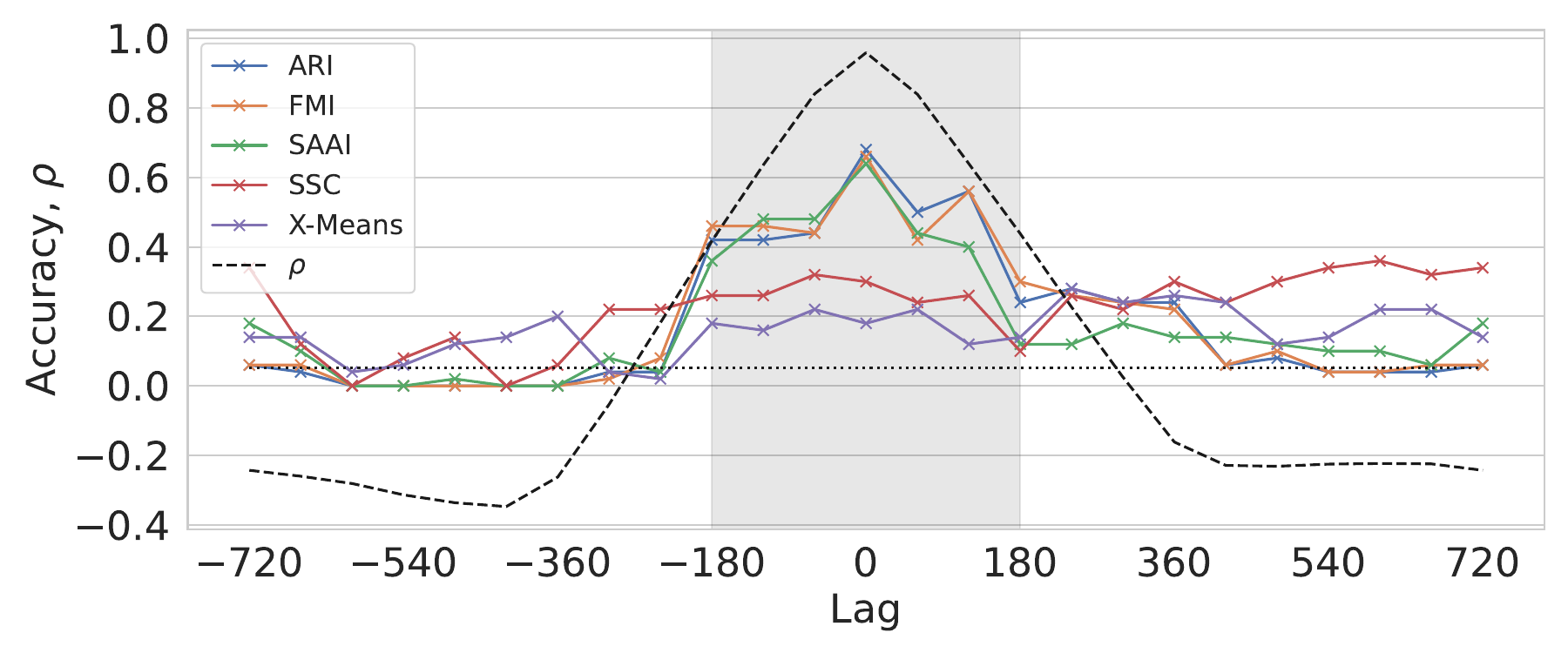}
    \caption{Accuracies for finding the correct value for $K$ while increasing the lag $l$ between the two variables of the time series from $-720$ minutes to $720$ minutes. The Pearson correlation Coefficient $\rho$ is shown as c black dahed line. The gray area between $l=-180$ and $l=180$ marks the sweet spot for applying SAAI as well as ARI and FMI. In this range, maximizing SAAI achieves superior accuracies compared to SSC. for $l=180$ the accuracy for X-Means is slighly higher ($0.14$ vs. $0.12$).}
    \label{fig:res:synth:shifting_lags}
\end{figure}

\paragraph{Lagged Variables}
In the experiments described above, the time series variables were all highly correlated. In Section~\ref{sec:method:saai} we derived SAAI for "temporally aligned anomalies in similar measurements". To get an idea of "how similar" the signals of the multivariate time series need to be, we perform the following experiment: We fix $D=2$, select $r_{sync} \in [0.5, 1]$ and $K=4$ new classes uniformly at random on every run. We modify the correlation between the variables of the $2D$ time series by increasing the lag $l$ between the first and second dimensions in steps of $60$ minutes, from $l=-720$ ($-0.5$ day) to $l=720$ ($+0.5$ day). Again, we measure the accuracy of finding the true value $K$ by maximizing SAAI, SSC, ARI, and FMI and by applying X-Means. The results are shown in Figure \ref{fig:res:synth:shifting_lags}. The black dashed line shows the Pearson correlation coefficient $\rho$ for the two variables of the time series. 
The baseline, based on random guessing, for finding the correct value of $K$ for $2 \leq k < 20$ is $p=\frac{1}{19}$ and shown as a black dotted line.

For $-180 \leq l \leq 180$, the shaded gray in Figure \ref{fig:res:synth:shifting_lags}, SAAI achieves superior results than SSC and is again almost on par with ARI and FMI. For X-Means, the accuracy of $0.14$ at $l=180$ is slightly higher than $0.12$ for SAAI. The area of $-180 \leq l \leq 180$ corresponds to a correlation of $\rho > 0.43$ and marks the sweet spot for applying SAAI. For $l \geq 360$ maximizing SSC shows better or equal results compared to all other methods.

\begin{figure}[ht]
    \centering
    \includegraphics[width=.95\columnwidth]{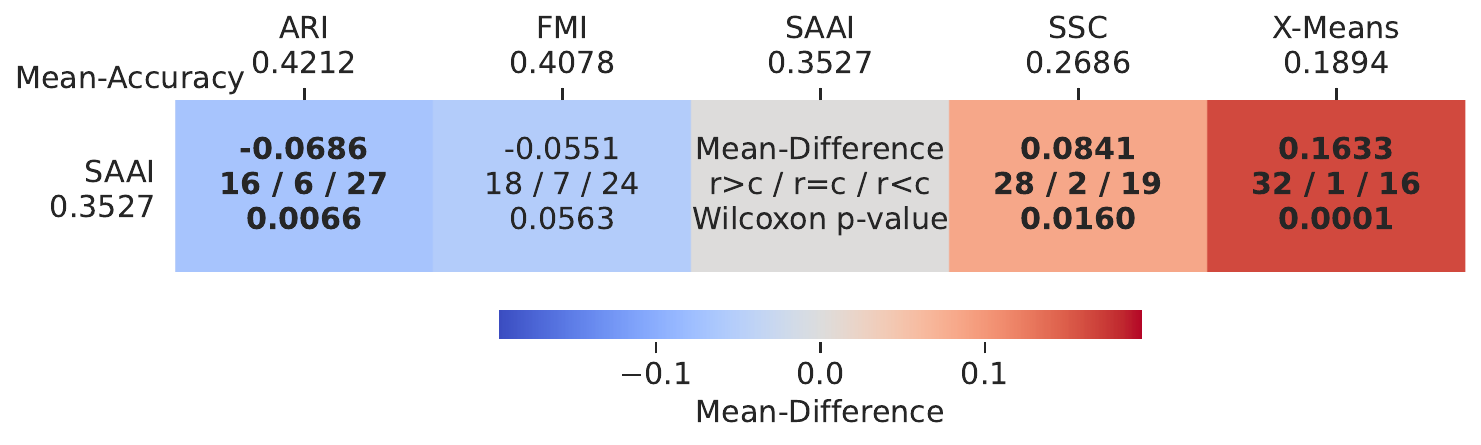}
    \caption{Multi-Comparison Matrix summarizing the results of the experiments in Section \ref{sec:res:exp1}.}
    \label{fig:res:synth:mcm}
\end{figure}

\paragraph{Summary}
The Multi Comparison Matrix (MCM) \cite{Ismail2023} shown in Figure \ref{fig:res:synth:mcm} summarizes the results presented before. It shows the Mean Accuracy for the task of finding the correct value for K of SAAI compared to its competitors. Each cell of the MCM shows the difference in mean accuracy between SAAI and the respective competitor in the top row. The middle row contains the number of wins, ties and losses, where "win" means, that SAAI achieved a higher accuracy than the respective competitor in one experiment. The bottom row shows the p-value of the Wilcoxon signed rank test \cite{Wilcoxon1945}, which is a non-parametric test used to compare paired  samples, without assuming normal distribution of the data. The tested null hypothesis ($H_0$) is, that the distribution of differences between the paired observations are symmetric around zero. For our case, we can formulate $H_0$ as: There is no difference in the central tendency between pairs of methods for finding the true value for K.
The values in a cell are printed in bold, if the p-value is below $0.05$ and hence $H_0$ is rejected, indicating statistical significance.

As expected, maximizing SAAI is outperformed by maximizing ARI and FMI, which had access to the ground truth labels. Interestingly does this advantage not lead to a significant improvement for FMI. Maximizing SSC and X-Means perform statistically significantly worse than maximizing SAAI.

\begin{figure}[ht]
    \centering
    \includegraphics[width=.5\columnwidth]{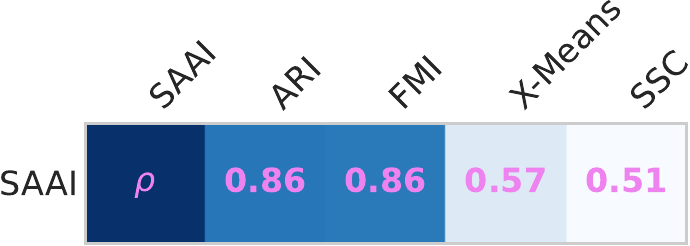}
    \caption{Correlation of the results of SAAI with the competing methods.}
    \label{fig:res:synth:correlation}
\end{figure}

The correlation plot shown in Figure \ref{fig:res:synth:correlation} shows the average correlation coefficient $\rho$ with respect to the experimental results between SAAI and its competitors. Maximizing SAAI shows high to very high correlation \cite{Mukaka2012} with maximizing ARI and FMI, but only moderate to low correlation \cite{Mukaka2012} with X-Means and SSC.

\hfill

\subsection{Real Greenhouse Temperature Data}
\label{sec:res:real}
In this experiment, we demonstrate the effectiveness of SAAI when working with real, unlabeled data. For this purpose, we use the ICS temperature measurements included in the \textit{edeniss2020} dataset \cite{Rewicki2024a}. This time series consists of 38 temperature measurements from the EDEN ISS research greenhouse. We follow the approach in \cite{Rewicki2024} and find anomalous sub-sequences using the algorithms Maximally Divergent Intervals (MDI) \cite{Barz2018} and Discord Aware Matrix Profile (DAMP) \cite{Lu2022} and cluster the anomalous subsequences using K-Means clustering after removing high frequent noise using moving average smoothing with window size $w=5$. We initialize K-means using kmeans++ initialization \cite{Arthur2007}. Since the anomalous sequences found by MDI and DAMP vary in length, we use DTW as distance metric. We run K-Means with increasing number of clusters $k$ from $2$ to $20$ and determine the metric-specific optimal number of clusters by maximizing SSC and SAAI, respectively. The optimal clusterings for SAAI and SSC are shown in Figure {\ref{fig:ics:saai_vs_ssc}}. The detailed results can be found in Figure \ref{fig:res:edeniss:scores_by_k} in the Appendix.

\begin{figure}[ht]
    \centering
    \begin{subfigure}[b]{.27\columnwidth}
        \centering
        \includegraphics[width=\columnwidth]{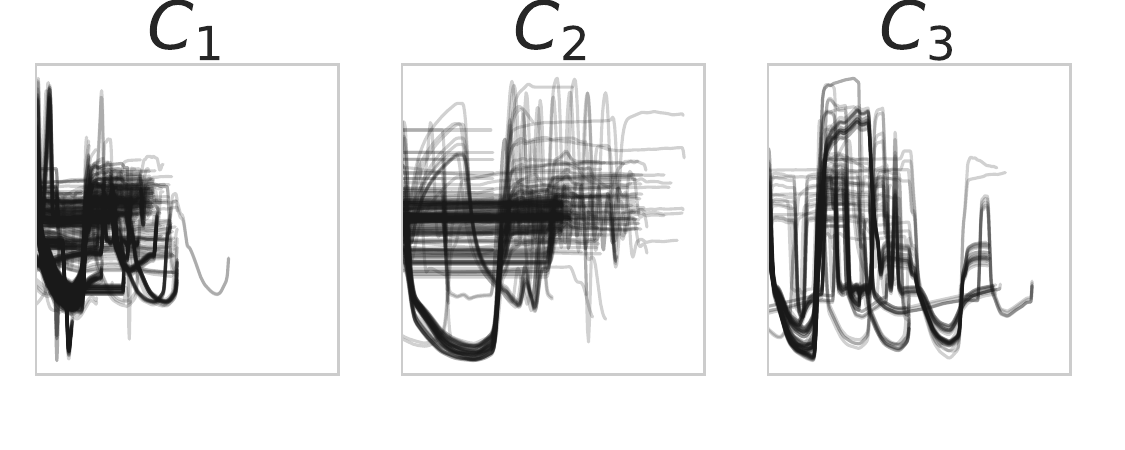}
        \caption{SSC}
        \label{fig:ics:ssc}
    \end{subfigure}
    \\
    \begin{subfigure}[b]{\columnwidth}
        \centering
        \includegraphics[width=\columnwidth]{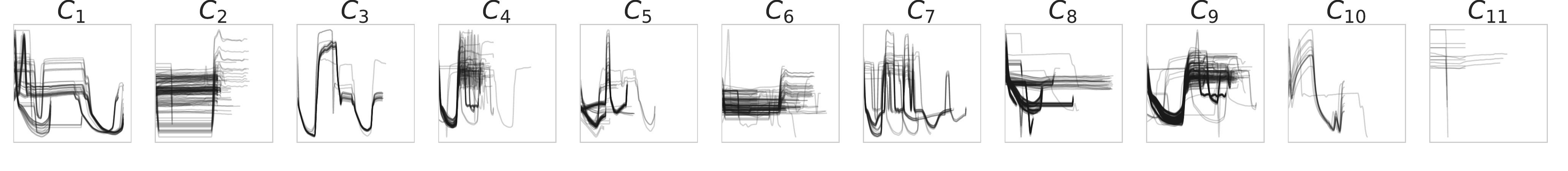}
        \caption{SAAI}
        \label{fig:ics:saai}
    \end{subfigure}
    \caption{Clustering solutions selected by maximizing (a) SSC, and 
    (b) SAAI. The results obtained by maximizing SAAI yield a better clustering in terms of visual interpretability and anomaly type determination.}
    \label{fig:ics:saai_vs_ssc}
\end{figure}

Maximizing SSC yields $3$ clusters, while maximizing SAAI yields $11$ clusters, which is much closer to the 10 anomaly types identified in \cite{Rewicki2024} for this time series. The cluster solution identified by maximizing SAAI is also easier to interpret. While we can hardly identify any anomaly types by visual inspection in the $3$-cluster solution returned by maximizing SSC, we can identify at least six anomaly types in the $11$-cluster solution found by maximizing SAAI. These anomaly types are \textit{Peak (short)} ($C_1$), \textit{Missing Night Phase} ($C_2$, $C_6$), \textit{Peak (long)} ($C_3$), \textit{Anomalous Day Phase} ($C_9$), \textit{Decreasing Peaks} ($C_{10}$) \textit{Near Flat Noisy or Flat signal} ($C_{11}$).

\subsection{Ablation Study}
\label{sec:res:ablation}
\begin{figure}
    \centering
    \includegraphics[width=.95\columnwidth]{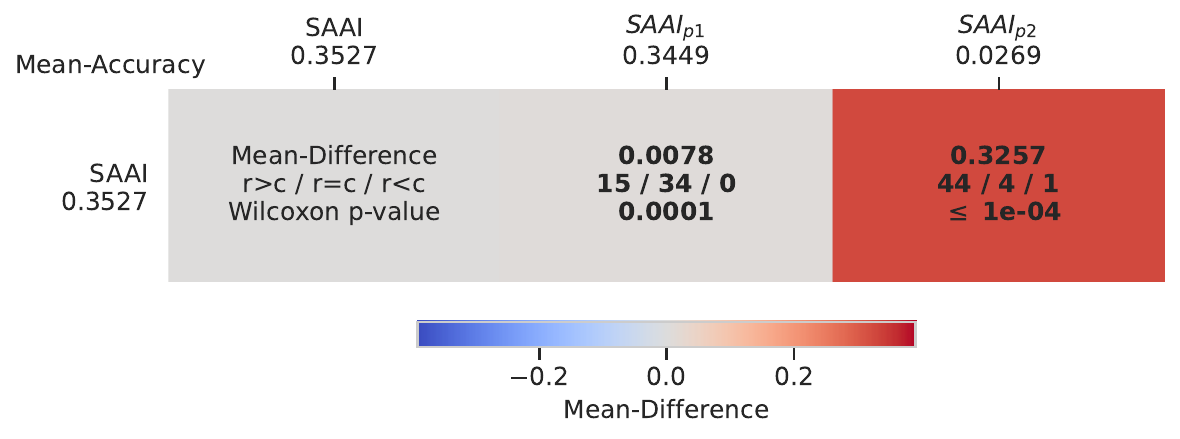}
    \caption{MCM comparing the proposed SAAI with versions using only the first or second penalty term.}
    \label{fig:res:ablation}
\end{figure}

In our ablation study, we evaluate the contribution of the penalty terms $\frac{1}{K}$ and $\frac{n_\mathds{1}}{K}$ in Equation (\ref{eq:saai}). We perform the same experiments as described in Section \ref{sec:res:exp1}. The results are summarized in Figure \ref{fig:res:ablation}. SAAI refers to the SAAI given in Definition \ref{def:saai}, while $\text{SAAI}_{p1}$ and $\text{SAAI}_{p2}$ refer to the SAAI with only the first and second penalty terms, respectively:

\begin{equation}
\begin{aligned}
    SAAI_{p1} &:= \lambda \frac{|A^*_S|}{|A_S|} + (1-\lambda)\frac{K-1}{K} \text{ , } \\
    SAAI_{p2} &:= \lambda \frac{|A^*_S|}{|A_S|} + (1-\lambda)\frac{K-n_\mathds{1}}{K} \text{ , }
\end{aligned}
\end{equation}
with $\lambda \in [0,1]$.

While the effect of penalizing pseudo-clusters through the second penalty term $\frac{n_\mathds{1}}{K}$ is smaller compared to penalizing small values for $K$, both terms add significant improvement on the overall accuracy.

\section{Discussion}
\label{sec:discussion}

Through our experiments in Section \ref{sec:res:exp1}, we have shown that maximizing SAAI outperforms maximizing SSC, as proposed by \cite{ShahapureNicholas2020,Zhou2014}, as well as X-Means, proposed by \cite{Pelleg2000} on the task of finding the true number of anomaly classes $K$ in multivariate time series consisting of sufficiently similar measurements.
Maximizing SAAI improves mean accuracy significantly over SAAI by $0.09$ and over X-Means by alomst $0.17$. The difference in mean accuracy of SAAI and FMI however is not statistically significant. 
The relatively low scores across all methods are subject to all runs from the \textit{Lagged Variables} experiment being included in the evaluation.
SAAI also shows a high to very high correlation with those results obtained by maximizing ARI and FMI.  Our results are consistent with those of \cite{Raihan2023} that SSC is not suitable for finding the correct value for $K$ by maximizing SSC when working with raw time series. Our findings contradict the proposal of \cite{ShahapureNicholas2020} and \cite{Zhou2014}, however they did not evaluate their approaches on time series data. 
\begin{figure}
    \centering
    \includegraphics[width=.95\columnwidth]{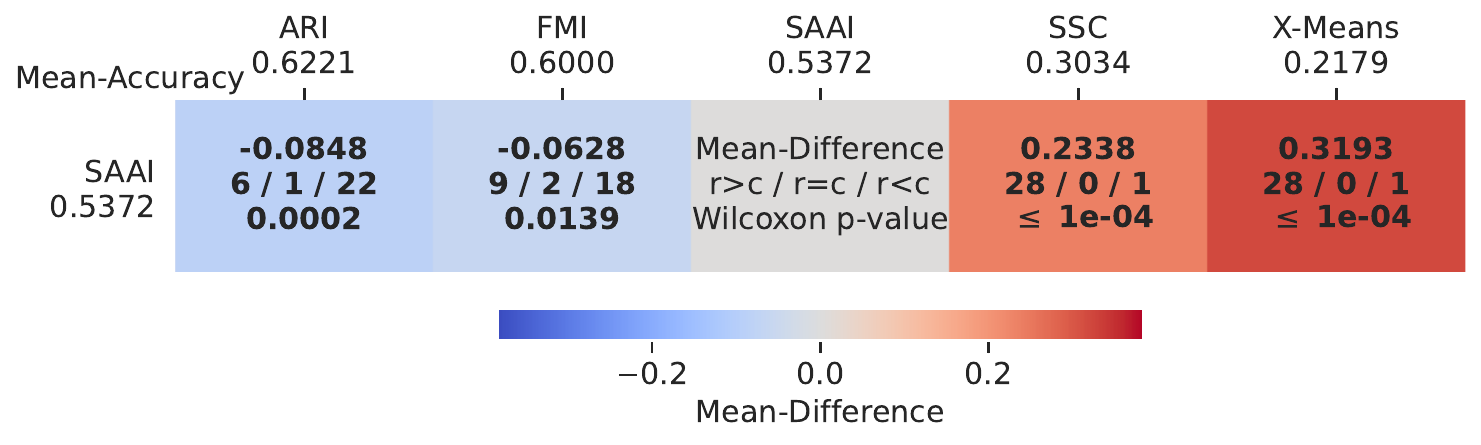}
    \caption{MCM comparing the results within the sweet spot of SAAI.}
    \label{fig:mcm:sweetspot}
\end{figure}
The \textit{Lagged Variables} experiments give an idea of how the rather vague notion of \textit{similar-enough} might be quantized. For correlation coefficients $\rho > 0.43$, maximizing SAAI gives an higher accuracy as maximizing SSC. However, for $l = 180$, which corresponds to $\rho = 0.43$, X-Means yields a slightly higher accuracy of $0.14$ as SAAI with $0.12$.
Thus, as a rule of thumb, we could say that SAAI is superior in finding the correct value for $K$ in similar measurements if their pairwise correlation coefficients $\rho$ satisfy $\rho \geq 0.5$. When applied to real sensor data, as done in Section \ref{sec:res:real}, we saw that the solution obtained by maximizing SAAI is easier to interpret compared to SSC. This is to be expected, since SAAI has already been shown to be superior in finding the ideal number of clusters. This result is also supported by those in \cite{Rewicki2024}, where for the same ICS temperature time series, 10 clusters (9 anomaly types and one false-positive cluster) were identified. 
Our ablation study showed that, although the first penalty term is more influential, both are needed, especially when the ratio of synchronized anomalies to all anomalies $\frac{|A_S|}{|A|}$ is low. The influence of the second term would increase as the range of possible values $k$ is increased, which would increase the likelihood of seeing pseudo-clusters.
However, SAAI has two shortcomings. Since SAAI only considers synchronized anomalies, as defined in Definition \ref{def:syncanom}, anomalies found in only one of the variables are not evaluated. Another drawback is the dependence on the similarity of the time-dependent signals. This limits the application of SAAI to these use cases, while SSC can be applied to arbitrary data.

\section{Conclusions}
\label{sec:conclusion}
In this paper, we propose SAAI, an unsupervised measure of anomaly cluster quality that incorporates prior knowledge about the multivariate time series by exploiting the similarity between individual signals. 
We demonstrate the effectiveness of SAAI by showing that maximizing SAAI outperforms maximizing SSC and X-Means on the task of finding the true number of anomaly classes $K$. 
Also, SAAI shows high correlation with results obtained from maximizing the external measures ARI and FMI.  
When applied to real, unlabeled data, the clustering result found by maximizing SAAI is easier to interpret compared to SSC. 
Our ablation study shows that all parts of the SAAI formula are necessary. 
However, SAAI has two major shortcomings: (1) it is only applicable to univariate anomalies found in multivariate time series consisting of reasonably similar signals, and (2) SAAI does not consider anomalies found in only a single variable (i.e., unaligned). 
Both shortcomings will be the subject of future research, as addressing (1) would allow extension to multivariate anomalies, and including unaligned anomalies by addressing (2) will expand the range of valid use cases.

\appendix
\section{Appendix}
\label{sec:appendix}
\subsection{SAAI Algorithm \& Complexity Analysis}
\label{appendix:algo}
The algorithm for calculating the SAAI is shown in algorithm \ref{algo:saai}.
To determine the set of synchronized anomalies $A_S$, we have to compare all anomalous subsequences in different variables $\mathcal{S}^{(i)}_{a_i,b_i}, \mathcal{S}^{(j)}_{a_j,b_j}$ that overlap in time, i.e. $i < j \text{ and } b_j < a_i$ or $b_i > a_j$.
A sweep-line algorithm for calculating the SAAI is given in Algorithm~\ref{algo:saai}.
For each anomalous sequence, we create two events in lines \ref{algo:saai:createEventsStart}-\ref{algo:saai:createEventsEnd} with a complexity of $\mathcal{O}(n)$, where $n = |A|$.
The sorting of the $2n$ events in line \ref{algo:saai:sort} has a complexity of $\mathcal{O}(n \log n)$. 
The events are sorted by time and in case of ties by event, so that "END" events are sorted before "START" events. 
Each interval is added to (line \ref{algo:saai:addE}) and removed from (line \ref{algo:saai:delE}) the active intervals $S$ once, resulting in $\mathcal{O}(n)$ insertions and deletions from $S$. 
Before an interval is added, it is compared to all active intervals. 
Since all intervals can be active at the same time, the maximum number of comparisons is $\binom{n}{2}$ in the worst case, which is in $\mathcal{O}(n^2)$.
However, this worst case occurs only if $n$ is close to the dimension of the time series $D$ and all anomalous subsequences are synchronized. Typically, we have $D \ll n$ when clustering anomalous subsequences in multivariate time series.
This gives a complexity of $\mathcal{O}(n \log n)$ for the average case where $D \ll n$ and overlaps are sparse, and $\mathcal{O}(n^2)$ for the worst case where $D \approx n$ and many overlapping intervals.

\begin{algorithm}
\caption{SAAI Algorithm}
\label{algo:saai}
    \textbf{Input}:
    \begin{itemize}
        \item[] $A$: The set of anomalies
    \item[]  $K$: Number of clusters
    \item[]  $n_\mathds{1}$: Number of pseudo-clusters, i.e. number of clusters with only one element
    \item[]  $\theta_{s}$: Degree of alignment between subsequences to be considered synchronous.
    \item[]  $\lambda$: Parameter to weight main and penalty term.
    \end{itemize}
    
    \textbf{Output}: saai
\begin{algorithmic}[1]
    \STATE $A_\mathcal{S}, A^{*}_\mathcal{S} \gets \varnothing, \varnothing$
    \STATE $E, S \gets [\text{ }], [\text{ }] $ 
    \STATE $i \gets 0 $
    \FOR {$S_{a,b} \in A$}\label{algo:saai:createEventsStart}
        \STATE append $\lbrace{("START", a, i, b)\rbrace}$ to $E$ 
        \STATE append $\lbrace{("END", b, i)}\rbrace$ to $E$ 
        \STATE $i \gets i+1$
    \ENDFOR\label{algo:saai:createEventsEnd}
    \STATE $E \gets sort(E, \text{time}, \text{type})$ \label{algo:saai:sort} 
    \FOR {all events $e \in E$}
        \IF{$e.type = "START"$}
            \FOR{active intervals $s \in S$}
                \STATE $v = \omega(e.time, e.end, s.time, s.end)$
                \IF{$v \geq \theta$ }
                    \STATE $A_\mathcal{S} = A_\mathcal{S} \cup \lbrace{(A[e.id], A[s.id])\rbrace}$
                    \IF{$c(A[e.id]) = c(A[s.id])$}  
                        \STATE $A^{*}_\mathcal{S} = A^{*}_\mathcal{S} \cup \lbrace{(A[e.id], A[s.id])\rbrace}$ 
                    \ENDIF
                \ENDIF
            \ENDFOR
            \STATE append $e$ to $S$ \label{algo:saai:addE}
        \ELSE
            \STATE delete $e$ from $S$ 
            \label{algo:saai:delE}
        \ENDIF
    \ENDFOR
    
    \STATE $saai \gets \lambda \frac{|A^*_S|}{|A_S|} + (1-\lambda)\frac{K-1-n_\mathds{1}}{K}$
    \STATE \RETURN $saai$
\end{algorithmic}
\end{algorithm}

\subsection{Selecting $\lambda$}
\begin{figure}[hb]
    \centering
    \includegraphics[width=\columnwidth]{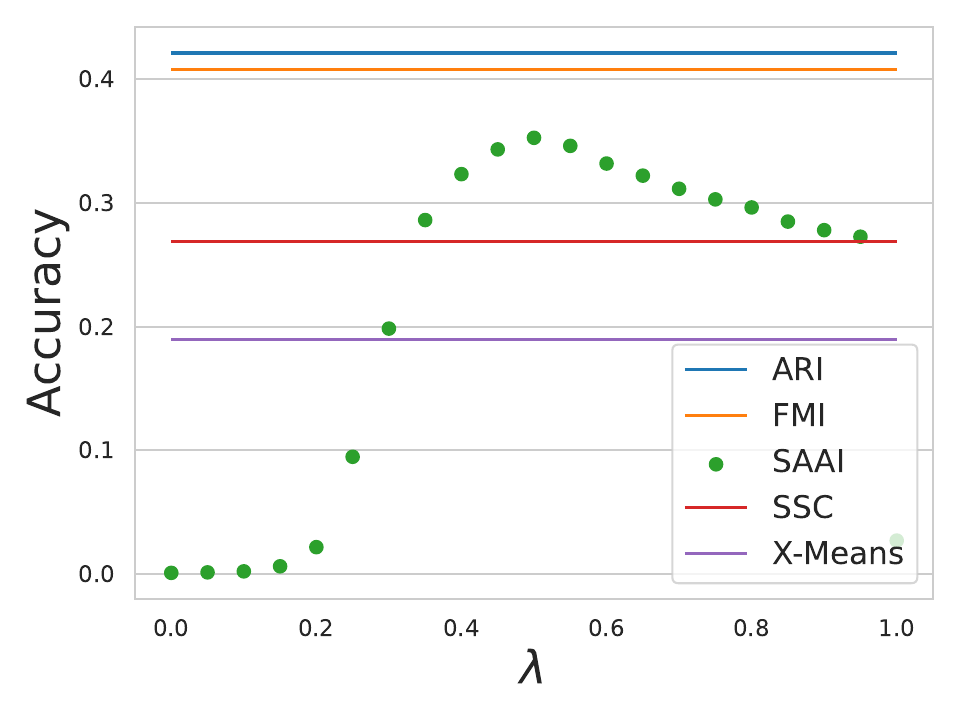}
    \caption{Accuracy of SAAI for increasing $\lambda$ compared to ARI, FMI, SSC and X-Means aggregated over all experimental results on the synthetic greenhouse temperature data.}
    \label{fig:res:synth:lambda}
\end{figure}

The parameter $\lambda$ in Equation (\ref{eq:saai}) determines the weight of the main term over the regularizing term. A value of $\lambda = 1$ would evaluate only the main term and ignore the number of clusters and pseudo-clusters in the solution found. On the contrary, a value of $\lambda=0$ would evaluate only the number of clusters and pseudo-clusters and ignore the synchronicity of anomalies.
Figure \ref{fig:res:synth:lambda} plots the mean accuracy for finding the true number of classes over all experiments on the synthetic greenhouse temperature data presented in Section \ref{sec:results} for increasing $\lambda$.
As can be seen in Figure \ref{fig:res:synth:lambda}, weighting the main and the regularizing term equally gives the best result for this example. However, there may be situations where weighting the two terms differently makes sense, e.g. when there is no preference for a larger number of clusters.
When choosing a value for $\lambda$, care should be taken, especially when giving more weight to the regularizing term, as this can be more detrimental to the performance of SAAI than giving more weight to the main term.
In general, weighting both terms equally by setting $\lambda = 0.5$ is a good initial choice.

\subsection{Silhouette Score Implementations}
\begin{figure}
    \centering
    \includegraphics[width=.95\columnwidth]{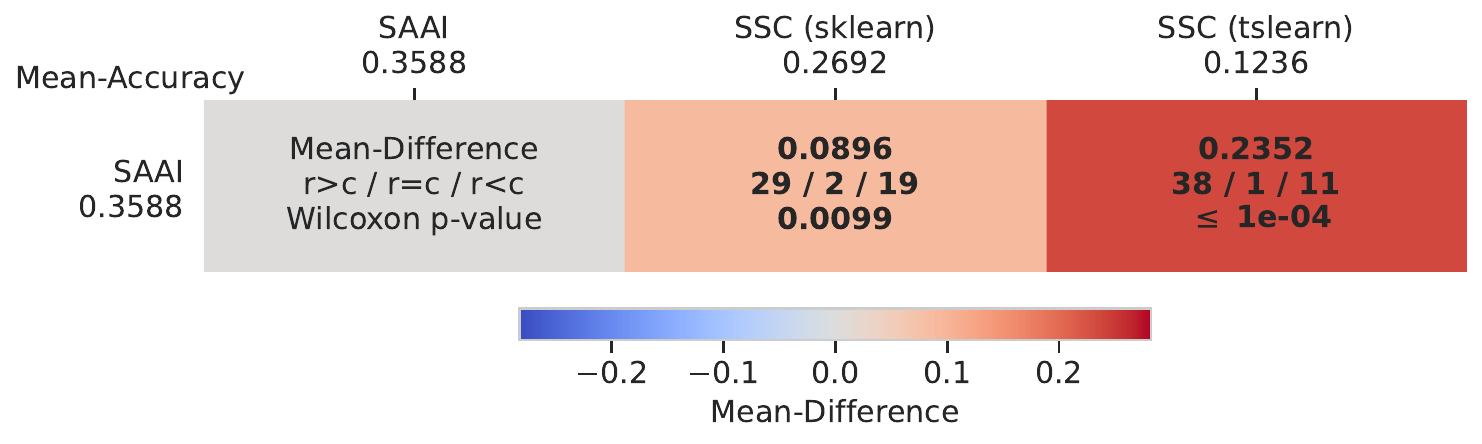}
    \caption{Multi-Comparison Matrix comparing the results obtained by using different implementations of the Silhouette Score.}
    \label{fig:res:ssc:mcm}
\end{figure}
While running our experiments on synthetic data, we found that the results for the Silhouette Score depend strongly on the implementation used to compute it. As shown in Figure \ref{fig:res:ssc:mcm}, the implementation in the \textit{tslearn} package \cite{TSLearn} gives significantly worse results than the implementation in \textit{scikit-learn}. This is surprising since \textit{tslearn} is a specialized package for time series analysis and supports the calculation of the Silhouette score for sequences of unequal length. However, for the sake of a fair comparison, we decided to use the \textit{scikit-learn} implementation in our main experiments, since the average accuracy of $0.2692$ is more than double that of $0.1236$ using the \textit{tslearn} implementation.

\subsection{SAAI and SSC results for edeniss2020 (ICS) dataset}
\begin{figure}[hb]
    \centering
    \includegraphics[width=\columnwidth]{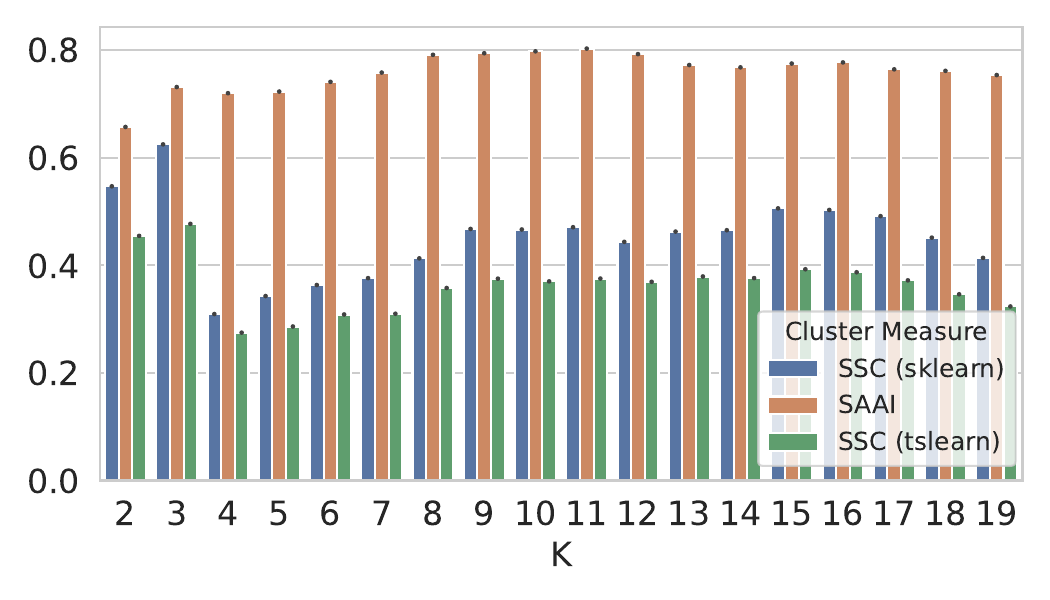}
    \caption{SAAI and SSC scores (tslearn and scikit-learn) for $1 < K < 20$}
    \label{fig:res:edeniss:scores_by_k}
\end{figure}

For the experiment on real ICS data from the EDEN ISS research greenhouse, we clustered the anomalous subsequences found by the MDI and DAMP algorithms with increasing number of clusters $1 < K < 20$. The results of SAAI and SSC are visualized in Figure \ref{fig:res:edeniss:scores_by_k}. Both SSC variants have their highest score at $K=3$, while SAAI has its maximum at $K=11$, which seems to be a more realistic value in this case. 

\subsection{SAAI Example (high res)}
\begin{figure}[ht]
    \centering
    \includegraphics[width=\columnwidth]{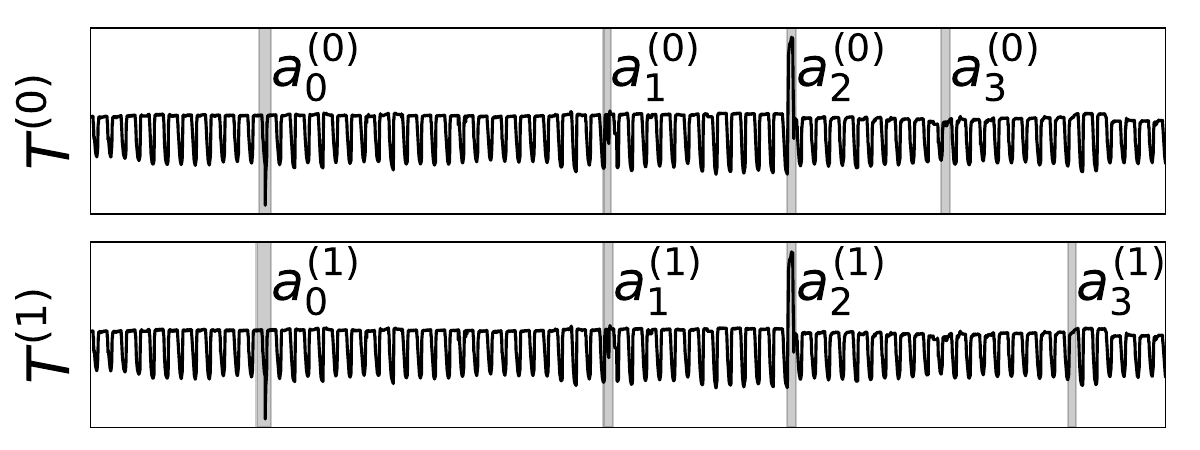}
    \caption{The detected anomalies $a^{(i)}_j$}
    \label{fig:saai:ex1:hr}
\end{figure}

\begin{figure}[ht]
    \begin{subfigure}{\columnwidth}
        \centering
        \includegraphics[width=\textwidth]{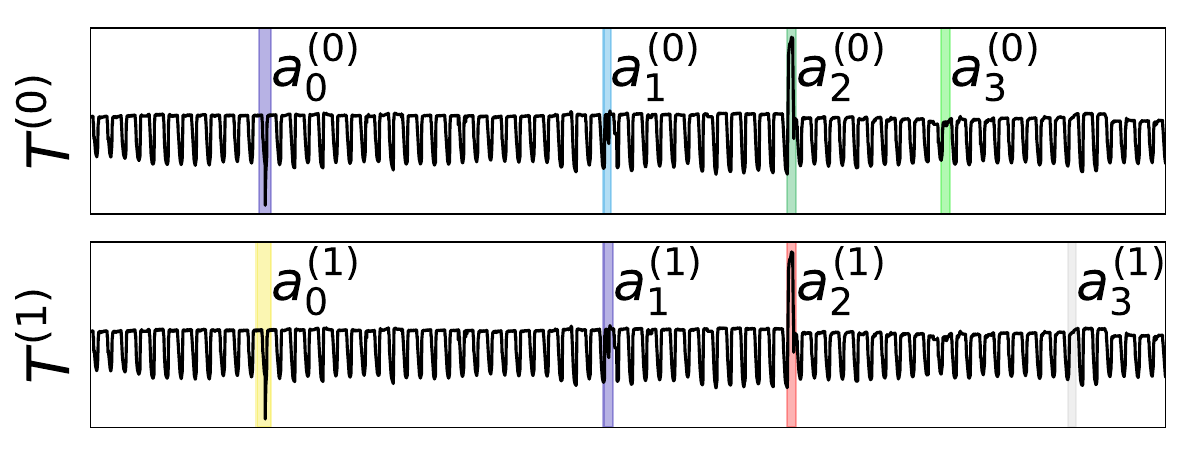}
        \caption{$SAAI = 0$}
        \label{fig:saai:ex6:hr}
    \end{subfigure}
\end{figure}
\begin{figure}[ht]\ContinuedFloat
    
    \begin{subfigure}{\columnwidth}
        \centering
        \includegraphics[width=\columnwidth]{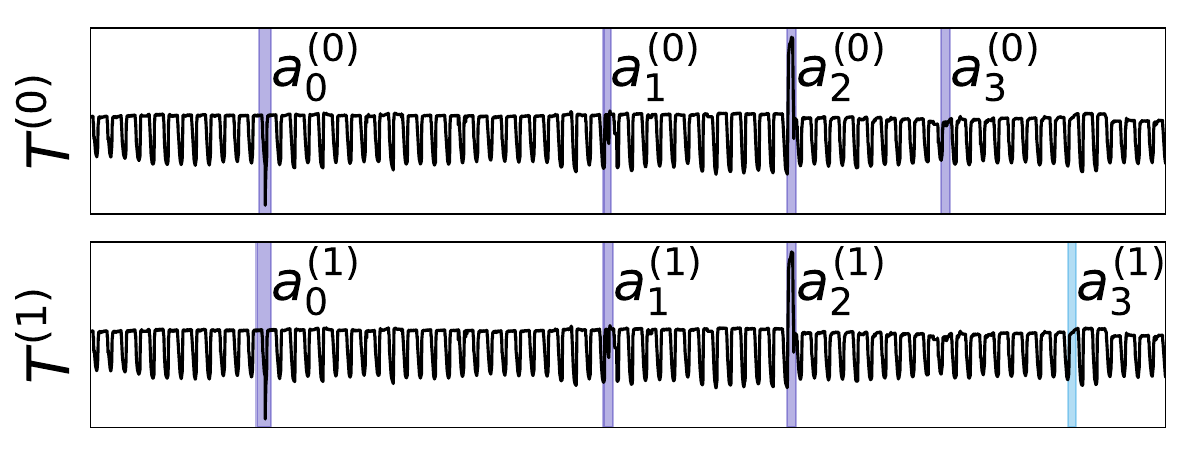}
        \caption{$SAAI = 0.5$}
        \label{fig:saai:ex2:hr}
    \end{subfigure}
    \begin{subfigure}{\columnwidth}
        \centering
        \includegraphics[width=\columnwidth]{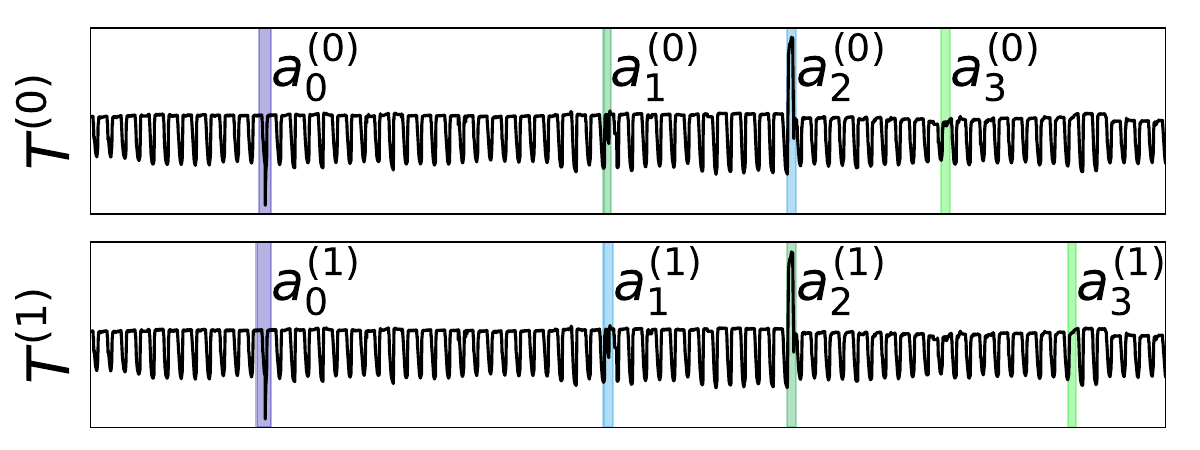}
        \caption{$SAAI = 0.541\bar{6}$}
        \label{fig:saai:ex5:hr}
    \end{subfigure}
    \begin{subfigure}{\columnwidth}
        \centering
        \includegraphics[width=\columnwidth]{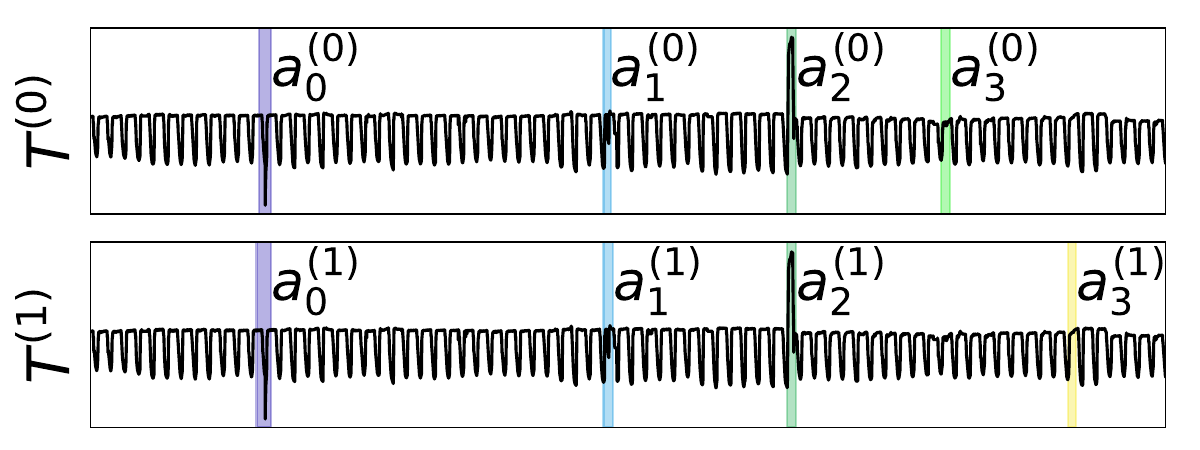}
        \caption{$SAAI = 0.7$}
        \label{fig:saai:ex4:hr}
    \end{subfigure}
    \begin{subfigure}{\columnwidth}
        \centering
        \includegraphics[width=\columnwidth]{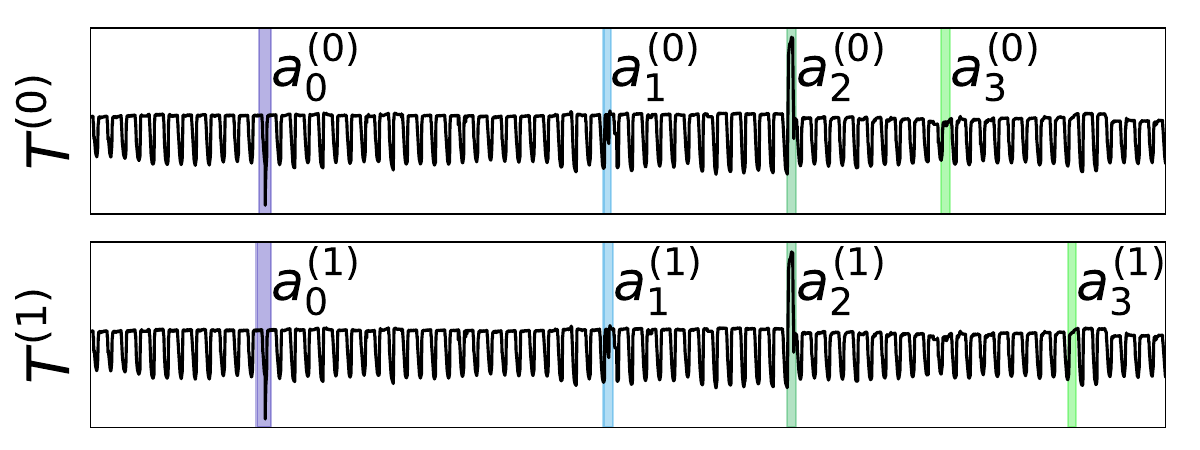}
        \caption{$SAAI = 0.875$}
        \label{fig:saai:ex3:hr}
    \end{subfigure}
    
    \caption{(a) - (e) different clustering solutions with increasing quality. Cluster assignment is coded by color. (a) Worst case: all but one cluster contain a single element, (b) all but one anomaly assigned to the same cluster, (c) synchronized anomalies not in the same cluster, (d) synchronized anomalies in separate clusters, single element clusters exist, (e) best case: synchronized anomalies in separate clusters, no single element cluster.}
    \label{fig:saai:examples}
\end{figure}

\end{document}